\documentclass{article}

\usepackage[final, nonatbib]{neurips_2019}

\usepackage[utf8]{inputenc} 
\usepackage[T1]{fontenc}    
\usepackage{hyperref}       
\usepackage{url}            
\usepackage{booktabs}       
\usepackage{amsfonts}       
\usepackage{nicefrac}       
\usepackage{microtype}      

\usepackage{graphicx}
\usepackage{subfigure}
\usepackage{amssymb}
\usepackage{mathtools}

\newcommand\numberthis{\addtocounter{equation}{1}\tag{\theequation}}
\newcommand*{\B}[1]{\mathbf{#1}}
\newcommand*{\C}[1]{\mathcal{#1}}
\newcommand*{\T}[1]{\textrm{#1}}

\newcommand\blfootnote[1]{%
  \begingroup
  \renewcommand\thefootnote{}\footnote{#1}%
  \addtocounter{footnote}{-1}%
  \endgroup
}

\usepackage{algorithm}
\usepackage{algorithmic}
\usepackage{hyperref}

\usepackage{wrapfig,lipsum,booktabs}

\title{Meta-Curvature}

\author{%
  Eunbyung Park \\
  Department of Computer Science\\
  University of North Carolina at Chapel Hill \\
  \texttt{eunbyung@cs.unc.edu} \\
  \And
  Junier B. Oliva \\
  Department of Computer Science \\
  University of North Carolina at Chapel Hill \\
  \texttt{joliva@cs.unc.edu} \\
}

\begin{document}

\maketitle

\begin{abstract}
We propose \textit{meta-curvature} (MC), a framework to learn curvature information for better generalization and fast model adaptation. MC expands on the model-agnostic meta-learner (MAML) by learning to transform the gradients in the inner optimization such that the transformed gradients achieve better generalization performance to a new task. For training large scale neural networks, we decompose the curvature matrix into smaller matrices in a novel scheme where we capture the dependencies of the model's parameters with a series of tensor products. We demonstrate the effects of our proposed method on several few-shot learning tasks and datasets. Without any task specific techniques and architectures, the proposed method achieves substantial improvement upon previous MAML variants and outperforms the recent state-of-the-art methods. Furthermore, we observe faster convergence rates of the meta-training process. Finally, we present an analysis that explains better generalization performance with the meta-trained curvature.
\end{abstract}

\section{Introduction}
\label{introduction}
\blfootnote{The code is available at \url{https://github.com/silverbottlep/meta_curvature}} Despite huge progress in artificial intelligence, the ability to quickly learn from few examples is still far short of that of a human. We are capable of utilizing prior knowledge from past experiences to efficiently learn new concepts or skills. With the goal of building machines with this capability, \textit{learning-to-learn} or \textit{meta-learning} has begun to emerge with promising results.

One notable example is model-agnostic meta-learning (MAML) \cite{Finn-icml-2017, Nichol-arxiv-2018}, which has shown its effectiveness on various few-shot learning tasks. It formalizes \textit{learning-to-learn} as meta objective function and optimizes it with respect to a model's initial parameters. Through the meta-training procedure, the resulting model's initial parameters become a very good prior representation and the model can quickly adapt to new tasks or skills through one or more gradient steps with a few examples. Although this end-to-end approach, using standard gradient descent as the inner optimization algorithm, was theoretically shown to approximate any learning algorithm \cite{Finn-iclr-2018}, recent studies indicate that the choice of the inner-loop optimization algorithm affects performance.  \cite{Li-arxiv-2017,Antoniou-arxiv-2018,Grant-iclr-2018}.

Given the sensitivity to the inner-loop optimization algorithm, second order optimization methods (or preconditioning the gradients) are worth considering. They have been extensively studied and have shown their practical benefits in terms of faster convergence rates \cite{Nocedal_2006}, an important aspect of few-shot learning. In addition, the problems of computational and spatial complexity for training deep networks can be effectively handled thanks to recent approximation techniques \cite{Martens-icml-2015,LeRoux-2008}. Nevertheless, there are issues with using second order methods in its current form as an inner loop optimizer in the meta-learning framework. First, they do not usually consider generalization performance. They compute local curvatures with training losses and move along the curvatures as far as possible. It can be very harmful, especially in the few-shot learning setup, because it can overfit easily and quickly. 

In this work, we propose to learn a curvature for better generalization and faster model adaptation in the meta-learning framework, we call \textit{meta-curvature}. The key intuition behind MAML is that there are some representations are broadly applicable to all tasks. In the same spirit, we hypothesize that there are some curvatures that are broadly applicable to many tasks. Curvatures are determined by the model's parameters, network architectures, loss functions, and training data. Assuming new tasks are distributed from the similar distribution as meta-training distribution, there may exist common curvatures that can be obtained through meta-training procedure. The resulting meta-curvatures, coupled with the simultaneously meta-trained model's initial parameters, will transform the gradients such that the updated model has better performance on new tasks with fewer gradient steps. In order to efficiently capture the dependencies between all gradient coordinates for large networks, we design a multilinear mapping consisting of a series of tensor-products to transform the gradients. It also considers layer specific structures, e.g. convolutional layers, to effectively reflects our inductive bias. In addition, meta-curvature can be easily implemented (simply transform the gradients right before passing through the optimizers) and can be plugged into existing meta-learning frameworks like MAML without additional, burdensome higher-order gradients. 

We demonstrate the effectiveness of our proposed method on the few-shot learning tasks done by \cite{matchingnet,Ravi-iclr-2017,Finn-icml-2017}. We evaluated our methods on few-shot regression and few-shot classification tasks over Omniglot \cite{Lake1332}, miniImagenet \cite{matchingnet}, and tieredImagnet \cite{ren-iclr18} datasets. Experimental results show significant improvements on other MAML variants on all few-shot learning tasks. 
In addition, MC's simple gradient transformation outperformed other more complicated state-of-the-art methods that include additional bells and whistles.

\vspace{-2mm}
\section{Background}
\label{background}

\vspace{-2mm}
\subsection{Tensor Algebra}
We review basics of tensor algebra that will be used to formalize the proposed method. We refer the reader to \cite{KoBa09} for a more comprehensive review. Throughout the paper, tensors are defined as multidimensional arrays and denoted by calligraphic letters, e.g. $N$th-order tensor, $\mathcal{X} \in \mathbb{R}^{I_1 \times I_2 \times \cdots \times I_N}$. Matrices are second-order tensors and denoted by boldface uppercase, e.g. $\mathbf{X} \in \mathbb{R}^{I_1 \times I_2}$.

\textbf{Fibers:} Fibers are a higher-order generalization of matrix rows and columns. A matrix column is a mode-$1$ fiber and a matrix row is a mode-$2$ fiber. The mode-1 fibers of a third order tensor $\mathcal{X}$ are denoted as $\mathcal{X}_{:,j,k}$, where a colon is used to denote all elements of a mode.

\textbf{Tensor unfolding:} Also known as \textit{flattening (reshaping)} or \textit{matricization}, is the operation of arranging the elements of an higher-order tensors into a matrix. The mode-$n$ unfolding of a $N$th-order tensor $\mathcal{X} \in \mathbb{R}^{I_1 \times I_2 \times \cdots \times I_N}$, arranges the mode-n fibers to be the columns of the matrix, denoted by $\mathcal{X}_{[n]} \in \mathbb{R}^{I_n \times I_M}$, where $I_M=\prod_{k \neq n} I_k$. The elements of the tensor, $ \mathcal{X}_{i_1,i_2,\dots,i_N}$ are mapped to $\mathcal{X}_{[n]i_n,j}$, where $j = 1 + \sum_{k \neq n, k=1}^{N} (i_k-1)J_k$, with $J_k=\prod_{m=1,m \neq n}^{k-1}I_m$.

\textbf{\textit{n}-mode product:} It defines the product between tensors and matrices. The $n$-mode product of a tensor $\mathcal{X} \in \mathbb{R}^{I_1 \times I_2 \times \cdots \times I_N}$ with a matrix $\mathbf{M} \in \mathbb{R}^{J \times I_n}$ is denoted by $\mathcal{X} \times_n \mathbf{M}$ and computed as
\begin{equation}
(\mathcal{X} \times_n \mathbf{M})_{i_1,\dots,i_{n-1},j,i_{n+1},\dots,i_N} = \sum_{i_n=1}^{I_n} \mathcal{X}_{i_1,i_2,\dots,i_N} \mathbf{M}_{j,i_n} .
\end{equation}
More concisely, it can be written as $(\mathcal{X} \times_n \mathbf{M})_{[n]} = \mathbf{M}\mathcal{X}_{[n]} \in \mathbb{R}^{I_1 \times \cdots \times I_{n-1} \times J \times I_{n+1} \times \cdots \times I_N}$. Despite cumbersome notation, it is simply $n$-mode unfolding (reshaping) followed by matrix multiplication.

\vspace{-2mm}
\subsection{Model-Agnostic Meta-Learning (MAML)}
MAML aims to find a transferable initialization (a prior representation) of any model such that the model can adapt quickly from the initialization and produce good generalization performance on new tasks. The meta-objective is defined as validation performance after one or few step gradient updates from the model's initial parameters. By using gradient descent algorithms to optimize the meta-objective, its training algorithm usually takes the form of nested gradient updates: inner updates for model adaptation to a task and outer-updates for the model's initialization parameters. Formally, 
\begin{equation}
\label{eq:maml}
\min_\theta \mathbb{E}_{\tau_i}[\mathcal{L}_\textrm{val}^{\tau_i} \big( \underbrace{\theta - \alpha \nabla \mathcal{L}_{\textrm{tr}}^{\tau_i}(\theta)}_{\textrm{inner udpate}} \big)],
\end{equation}
where $\mathcal{L}_\textrm{val}^{\tau_i}(\cdot)$ denotes a loss function for a validation set of a task $\tau_i$, and $\mathcal{L}_\textrm{tr}^{\tau_i}(\cdot)$ for a training set, or $\mathcal{L}_\textrm{tr}(\cdot)$ for brevity. The inner update is defined as a standard gradient descent with fixed learning rate $\alpha$. For conciseness, we assume as single adaptation step, but it can be easily extended to more steps. For more details, we refer to \cite{Finn-icml-2017}. Several variations of inner update rules were suggested. Meta-SGD \cite{Li-arxiv-2017} suggested coordinate-wise learning rates, $\theta - \alpha \circ \nabla \mathcal{L}_\textrm{tr}$, where $\alpha$ is the learnable parameters and $\circ$ is element wise product. Recently, \cite{Antoniou-arxiv-2018} proposed a learnable learning rate per each layers for more flexible model adaptation. To alleviate computational complexity, \cite{Nichol-arxiv-2018} suggested an algorithm that do not require higher order gradients.

\subsection{Second order optimization}
The biggest motivation of second order methods is that first-order optimization such as standard gradient descent performs poorly if the Hessian of a loss function is ill-conditioned, e.g. a long narrow valley loss surface. There are a plethora of works that try to accelerate gradient descent by considering local curvatures. Most notably, the update rules of Newton's method can be written as $\theta -  \alpha \mathbf{H}^{-1}\nabla \mathcal{L}_\textrm{tr}$, with Hessian matrix $\mathbf{H}$ and a step size $\alpha$ \cite{Nocedal_2006}. Every step, it minimizes a local quadratic approximation of a loss function, and the local curvature is encoded in the Hessian matrix. Another promising approach, especially in neural network literature, is natural gradient descent \cite{Amari}. It finds a steepest descent direction in distribution space rather than parameter space by measuring KL-divergence as a distance metric. Similar to Newton's method, it preconditions the gradient with the Fisher information matrix and a common update rule is $\theta - \alpha \mathbf{F}^{-1} \nabla \mathcal{L}_\textrm{tr}$. In order to mitigate computational and spatial issues for large scale problems, several approximation techniques has been proposed, such as online update methods \cite{Nocedal_2006, LeRoux-2008}, Kronecker-factored approximations \cite{Martens-icml-2015}, and diagonal approximations of second order matrices \cite{rmsprop,adam,adagrad}.

\section{Meta-Curvature}
\label{meta_curvature}

We propose to learn a curvature along with the model's initial parameters simultaneously via the meta-learning process. The goal is that the meta-learned curvature works collaboratively with the meta-learned model's initial parameters to produce good generalization performance on new tasks with fewer gradient steps. In this work, we focus on learning a meta-curvature and its efficient forms to scale large networks. We follow the meta-training algorithms suggested in \cite{Finn-icml-2017} and the proposed method can be easily plugged in. 

\subsection{Motivation}
We begin with the hypothesis that there are broadly applicable curvatures to many tasks. In training a neural network with a loss function, local curvatures are determined by the model’s parameters, the network architecture, the loss function, and training data. Since new tasks are sampled from the same or similar distributions and all other factors are fixed, it is intuitive idea that there may exist some curvatures found via meta-training that can be effectively applied to the new tasks. Throughout the meta-training, we can observe how the gradients affect the validation performance and use those experiences to learn how to transform or correct the gradient from the new task.

We take a learning approach because existing curvature estimations do not consider generalization performance, e.g. Hessian and the Fisher-information matrix. The local curvatures are approximated with only current training data and loss functions. Therefore, these methods may end up converging fast to a poor local minimum. This is especially true when we have few training examples.


\begin{figure*}[t]
\begin{center}
\centerline{\includegraphics[width=\linewidth]{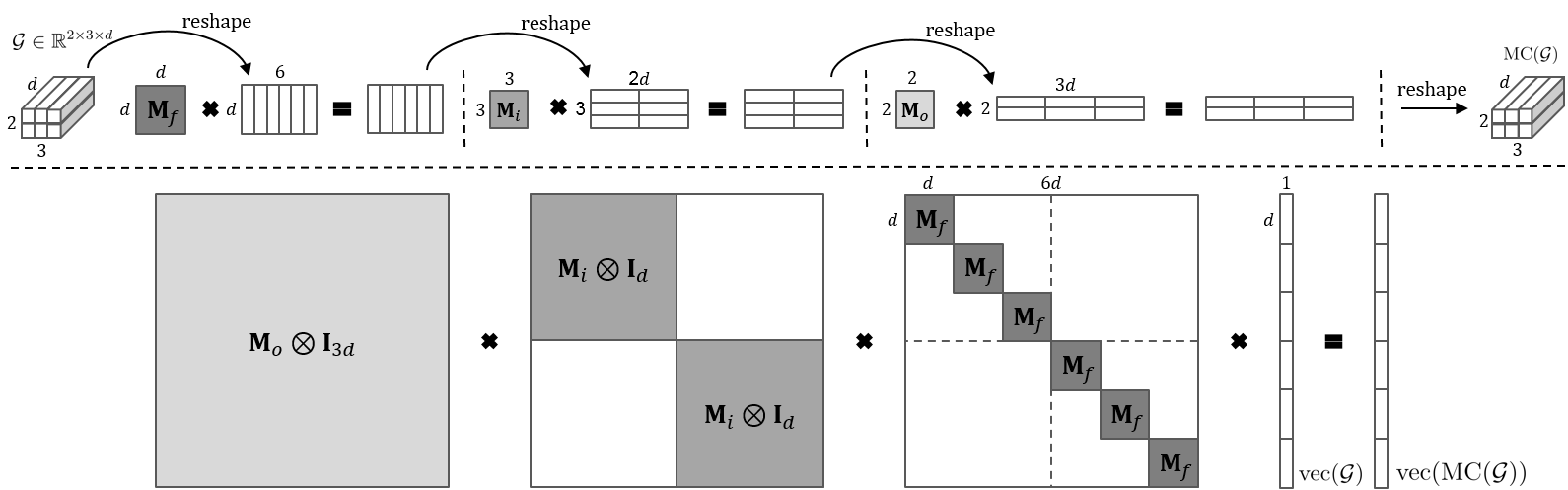}}
\caption{An example of meta-curvature computational illustration with $\mathcal{G} \in \mathbb{R}^{2 \times 3 \times d}$. Top: tensor algebra view, Bottom: matrix-vector product view.}
\label{fig_meta_curvature}
\end{center}
\vskip -0.3in
\end{figure*}

\subsection{Method}
First, we present a simple and efficient form of the meta-curvature computation through the lens of tensor algebra. Then, we present a matrix-vector product view to provide intuitive idea of the connection to the second order matrices. Lastly, we discuss the relationships to other methods.

\subsubsection{Tensor product view}
We consider neural networks as our models. With a slight abuse of notation, let the model's parameters $\mathcal{W}^l \in \mathbb{R}^{C_{out}^l \times C_{in}^l \times d^l}$ and its gradients of loss function $\mathcal{G}^l \in \mathbb{R}^{C_{out}^l \times C_{in}^l \times d^l}$, at each layers $l$. To avoid cluttered notation, we will omit the superscript $l$. We choose superscripts and dimensions with 2D convolutional layers in mind, but the method can be easily extended to higher dimension convolutional layers or other layers that consists of higher dimension parameters. $C_{out}, C_{in}$, and $d$ are the number of output channels, the number of input channels, and the filter size  respectively. $d$ is height $\times$ width in convolutional layers and 1 in fully connected layers. We also define meta-curvature matrices, $\mathbf{M}_o \in \mathbb{R}^{C_{out} \times C_{out}}$, $\mathbf{M}_{i} \in \mathbb{R}^{C_{in} \times C_{in}}$, and $\mathbf{M}_f \in \mathbb{R}^{d \times d}$. Now a meta-curvature function takes a multidimensional tensor as an input and has all meta-curvature matrices as learnable parameters:
\begin{equation}
\label{eq:mc_tensor}
\textrm{MC}(\mathcal{G}) = \mathcal{G} \times_3 \mathbf{M}_f \times_2 \mathbf{M}_{i} \times_1 \mathbf{M}_{o} .
\end{equation}
Figure~\ref{fig_meta_curvature} (top) shows an example of computational illustration with an input tensor $\mathcal{G} \in \mathbb{R}^{2 \times 3 \times d}$. First, it performs linear transformations for all $3$-mode fibers of $\mathcal{G}$. In other words, $\mathbf{M}_f$ captures the parameter dependencies between the elements within a $3$-mode fiber, e.g. all gradient elements in a channel of a convolutional filter. Secondly, the $2$-mode product models the dependencies between 3-mode fibers computed from the previous stage. All $3$-mode fibers are updated by linear combinations of other $3$-mode fibers belonging to the same output channel (linear combinations of $3$-mode fibers in a convolutional filter). Finally, the $1$-mode product is performed in order to model the dependencies between the gradients of all convolutional filters. Similarly, the gradients of all convolutional filters are updated by linear combinations of gradients of other convolutional filters. 

A useful property of $n$-mode products is the fact that the order of the multiplications is irrelevant for distinct modes in a series of multiplications. For example, $\mathcal{G} \times_3 \mathbf{M}_f \times_2 \mathbf{M}_{i} \times_1 \mathbf{M}_{o} = \mathcal{G} \times_1 \mathbf{M}_o \times_2 \mathbf{M}_{i} \times_3 \mathbf{M}_{f}$. Thus, the proposed method indeed examines the dependencies of the elements in the gradient all together.

\subsubsection{Matrix-vector product view}
We can also view the proposed meta-curvature computation as a matrix-vector product analogous to that from other second order methods. Note that this is for the purpose of intuitive illustration and we cannot compute or maintain this large matrices for large deep networks. We can expand the meta-curvature matrices as follows.
\begin{equation}
\label{eq:mc_tensor_expand}
\widehat{\mathbf{M}_o} = \mathbf{M}_o \otimes \mathbf{I}_{C_{in}} \otimes \mathbf{I}_{d}, \quad \widehat{\mathbf{M}_i} = \mathbf{I}_{C_{out}} \otimes \mathbf{M}_i  \otimes \mathbf{I}_{d}, \quad
\widehat{\mathbf{M}_f} = \mathbf{I}_{C_{out}} \otimes \mathbf{I}_{C_{in}} \otimes \mathbf{M}_f,
\end{equation}
where $\otimes$ is the Kronecker product, $\mathbf{I}_k$ is $k$ dimensional identity matrix, and the three expanded matrices are all same size $\widehat{\mathbf{M}_o}, \widehat{\mathbf{M}_i}, \widehat{\mathbf{M}_f} \in \mathbb{R}^{C_{out}C_{in}d \times C_{out}C_{in}d}$. Now we can transform the gradients with the meta-curvature as
\begin{equation}
\textrm{vec}(\textrm{MC}(\mathcal{G}))= \mathbf{M}_{mc}\textrm{vec}(\mathcal{G}),
\end{equation}
where $\mathbf{M}_{mc}= \widehat{\mathbf{M}_o} \widehat{\mathbf{M}_i} \widehat{\mathbf{M}_f}$. The expanded matrices satisfy commutative property, e.g. $\widehat{\mathbf{M}_o} \widehat{\mathbf{M}_i} \widehat{\mathbf{M}_f} = \widehat{\mathbf{M}_f} \widehat{\mathbf{M}_i} \widehat{\mathbf{M}_o}$, as shown in the previous section. Thus, $\mathbf{M}_{mc}$ models the dependencies of the model parameters all together. Note that we can also write $\mathbf{M}_{mc} = \mathbf{M}_o \otimes \mathbf{M}_i \otimes \mathbf{M}_f$, but this is non-commutative, $\mathbf{M}_o \otimes \mathbf{M}_i \otimes \mathbf{M}_f \neq \mathbf{M}_f \otimes \mathbf{M}_i \otimes \mathbf{M}_o$.

Figure~\ref{fig_meta_curvature} (bottom) shows a computational illustration. $\widehat{\mathbf{M}_f}\textrm{vec}(\mathcal{G})$, which is equivalent computation to $\mathcal{G} \times_3 \mathbf{M}_f$, can be interpreted as a giant matrix-vector multiplication with block diagonal matrix, where each block shares same meta-curvature matrix $\mathbf{M}_f$. It resembles the block diagonal approximation strategies in some second-order methods for training deep networks, but as we are interested in learning meta-curvature matrices, no approximation is involved. And matrix-vector product with $\widehat{\mathbf{M}_o}$ and $\widehat{\mathbf{M}_i}$ are used to capture inter-parameter dependencies and are computationally equivalent to $2$-mode and $3$-mode products of Eq. \ref{eq:mc_tensor}.

\subsubsection{Relationship to other methods}
Tucker decomposition \cite{KoBa09} decomposes a tensor into low rank cores with projection factors and aims to closely reconstruct the original tensor. We maintain full rank gradient tensors, however, and our main goal is to transform the gradients for better generalization. \cite{Kossaif-arxiv-2018} proposed to learn the projection factors in Tucker decomposition for fully connected layers in deep networks. Again, their goal was to find the low rank approximations of fully connected layers for saving computational and spatial cost.

Kronecker-factored Approximate Curvature (K-FAC) \cite{Martens-icml-2015,Grosse-icml-2016} approximates the Fisher matrix by the Kronecker product, e.g. $\B{F} \approx \B{A} \otimes \B{G}$, where $\B{A}$ is computed from the activation of input units and $\B{G}$ is computed from the gradient of output units. Its main goal is to approximate the Fisher such that matrix vector products between its inversion and the gradient can be computed efficiently. However, we found that maintaining $\B{A} \in \mathbb{R}^{C_{in}d \times C_{in}d}$ was quite expensive both computationally and spatially even for smaller networks. In addition, when we applied this factorization scheme to meta-curvature, it tends to easily overfit to meta-training set. On the contrary, we maintain two separated matrices, $\mathbf{M}_{i} \in \mathbb{R}^{C_{in} \times C_{in}}$ and $\mathbf{M}_f \in \mathbb{R}^{d \times d}$, which allows us to avoid overfitting and heavy computation. More importantly, we learn meta-curvature matrices to improve generalization instead of directly computing them from the activation and the gradient of training loss. Also, we do not require expensive matrix inversions.

\subsubsection{Meta-training}
We follow a typical meta-training algorithm and initialize all meta-curvature matrices as identity matrices so that the gradients do not change at the beginning. We used the ADAM \cite{adam} optimizer for the outer loop optimization and update the model's initial parameters and meta-curvatures simultaneously. We provide the details of algorithm in appendices.

\section{Analysis}
In this section, we will explore how a meta-trained matrix $\B{M}_{mc}$, or $\B{M}$ for brevity, can operate for better generalization. Let us take the gradient of meta-objective w.r.t $\B{M}$ for a task $\tau_i$. With the inner update rule $\theta^{\tau_i}(\B{M}) = \theta - \alpha  \B{M} \nabla_\theta \C{L}_\textrm{tr}^{\tau_i}(\theta)$, and by applying chain rule,
\begin{equation}
\nabla_{\mathbf{M}} \C{L}_\textrm{val}^{\tau_i}(\theta^{\tau_i}(\B{M})) = -\alpha \nabla_{\theta^{\tau_i}} \C{L}_\textrm{val}^{\tau_i} ( \theta^{\tau_i} ) \nabla_\theta \C{L}_\textrm{tr}^{\tau_i}(\theta)^\top, 
\end{equation}
where $\theta^{\tau_i}$ is the parameter for the task ${\tau_i}$ after the inner update.
It is the outer product between the gradients of validation loss and training loss. Note that there is a significant connection to the Fisher information matrix. For a task ${\tau_i}$, if we define the loss function as negative log likelihood, e.g. a supervised classification task $\mathcal{L}^{\tau_i} (\theta) = \mathbb{E}_{(x,y) \sim p({\tau_i})} [-\log_\theta p(y|x)$], then the empirical Fisher can be defined as $\mathbf{F} = \mathbb{E}_{(x,y) \sim p({\tau_i})} [\nabla_\theta \log_\theta p(y|x) \nabla_\theta \log_\theta p(y|x)^\top]$. There are three clear distinctions. First, the training and validation sets are treated separately in the meta-gradient $\nabla_{\B{M}} \C{L}_\textrm{val}^{\tau_i}$, while the empirical Fisher is computed with only training set (validation set is not available during training). Secondly, the gradient of the validation set is evaluated at new parameters $\theta^{\tau_i}$ after the inner update in the meta-gradient. Finally, the Fisher is positive semi-definite by construction, but it is not the case for the meta-gradient. This is an attractive property since it guarantees that the transformed gradient is always a descent direction. However, we mainly care about generalization performance in this work. Hence, we rather not force this property in this work, but leave it for future work.

Now let us consider what the meta-gradient can do for good generalization performance. Given a fixed point $\theta$ and a meta training set $\C{T} = \{ \tau_i \}$, standard gradient descent from an initialization $\B{M}$, gives the following update.
\setlength{\belowdisplayskip}{2pt} \setlength{\belowdisplayshortskip}{2pt}
\setlength{\abovedisplayskip}{2pt} \setlength{\abovedisplayshortskip}{2pt}
\begin{align}
\B{M}_\C{T} &= \B{M} - \beta \sum_{i=1}^{|\C{T}|} \nabla_\B{M} \C{L}_\textrm{val}^{\tau_i} (\theta^{\tau_i}(\B{M})) = \B{M} + \alpha \beta \sum_{i=1}^{|\C{T}|} \nabla_\theta \C{L}_\textrm{val}^{\tau_i} (\theta^{\tau_i}(\B{M})) \nabla_\theta \C{L}_\textrm{tr}^{\tau_i}(\theta)^\top,
\end{align}
where $\alpha$ and $\beta$ are fixed inner/outer learning rates respectively. Here, we assume a standard gradient descent for simplicity. But the argument extends to other advanced gradient algorithms, such as momentum and ADAM. 


We apply $\B{M}_\C{T}$ to the gradients of a new task, giving the transformed gradients
\setlength{\belowdisplayskip}{1pt} \setlength{\belowdisplayshortskip}{1pt}
\setlength{\abovedisplayskip}{1pt} \setlength{\abovedisplayshortskip}{1pt}
\begin{align}
\B{M}_\C{T} \nabla_\theta \C{L}_\T{tr}^{\tau_\T{new}}(\theta) &= \big( \B{M} + \alpha \beta \sum_{i=1}^{|\C{T}|} \nabla_\theta \C{L}_\T{val}^{\tau_i} (\theta^{\tau_i}) \nabla_\theta \C{L}_\T{tr}^{\tau_i}(\theta)^\top \big) \nabla_\theta \C{L}_\T{tr}^{\tau_\T{new}}(\theta) \\
\vspace{-2mm}
&= \B{M} \nabla_\theta \C{L}_\T{tr}^{\tau_\T{new}}(\theta) + \beta \sum_{i=1}^{|\C{T}|} \big( \nabla_\theta \C{L}_\T{tr}^{\tau_i}(\theta)^\top  \nabla_\theta \C{L}_\T{tr}^{\tau_\T{new}}(\theta) \big) \alpha \nabla_\theta \C{L}_\T{val}^{\tau_i} (\theta^{\tau_i}) \numberthis \label{eq:mc_new1} \\
\vspace{-2mm}
&= \B{M} \nabla_\theta \C{L}_\T{tr}^{\tau_\T{new}}(\theta) + \beta \sum_{i=1}^{|\C{T}|} \big( \underbrace{ \nabla_\theta \C{L}_\T{tr}^{\tau_i}(\theta)^\top  \nabla_\theta \C{L}_\T{tr}^{\tau_\T{new}}(\theta)}_{\T{A. Gradient similarity}} \big) \big( \underbrace{\alpha \nabla_\theta \C{L}_\T{val}^{\tau_i} (\theta) + \C{O}(\alpha^2)}_{\T{B. Taylor expansion}} \big). \numberthis \label{eq:mc_new2}
\end{align}

Given $\B{M}=\B{I}$, the second term in the R.H.S. of Eq. \ref{eq:mc_new2} can represent the final gradient direction for the new task. For Eq. \ref{eq:mc_new2}, we used the Taylor expansion of vector-valued function, $\nabla_\theta \C{L}_\T{val}^{\tau_i} (\theta^{\tau_i}) \approx \nabla_\theta \C{L}_\T{val}^{\tau_i} (\theta) +  \nabla_\theta^2 \C{L}_\T{val}^{\tau_i}(\theta) (\theta - \alpha \B{M} \nabla_\theta \C{L}_\T{tr}^{\tau_i}(\theta) - \theta)$.

The term A of Eq. \ref{eq:mc_new2} is the inner product between the gradients of meta-training losses and new test losses. We can simply interpret this as how similar the gradient directions between two different tasks. This has been explicitly used in continual learning or multi-task learning setup to consider task similarity \cite{Du-arxiv-2018,Lopez-nips2017,Riemer-iclr19}. When we have a loss function in the form of finite sums, this term can be also interpreted as a kernel similarity between the respective sets of gradients (see Eq. 4 of \cite{Muandet-nips2012}).

With the first term in B of Eq. \ref{eq:mc_new2}, we compute a linear combination of the gradients of validation losses from the meta-training set. Its weighting factors are computed based on the similarities between the tasks from the meta-training set and the new task as explained above. Therefore, we essentially perform a soft nearest neighbor voting to find the direction among the validation gradients from the meta-training set. Given the new task, the gradient may lead the model to overfit (or underfit). However, the proposed method will extract the knowledge from the past experiences and find the gradients that gave us good validation performance during the meta-training process.

\section{Related Work}
\label{related_work}
\textbf{Meta-learning:} Model-agnostic meta-learning (MAML) highlighted the importance of the model's initial parameters for better generalization \cite{Finn-iclr-2018} and there have been many extensions to improve the framework, e.g. for continuous adaptation \cite{Shedivat-iclr18}, better credit assignment \cite{promp}, and robustness \cite{Kim-nips2018}. In this work, we improve the inner update optimizers by learning a curvature for better generalization and fast model adaptation. Meta-SGD \cite{Li-arxiv-2017} suggests to learn coordinate-wise learning rates. We can interpret it as an diagonal approximation to meta-curvature in a similar vein to recent adaptive learning rates methods, such as \cite{rmsprop,adam,adagrad}, performing diagonal approximations of second-order matrices. Recently, \cite{Antoniou-arxiv-2018} suggested to learn layer-wise learning rates through the meta-training. However, both methods do not consider the dependencies between the parameters, which was crucial to provide more robust meta-training process and faster convergence. \cite{lee-icml-2018} also attempted to transform the gradients. They used simple binary mask applied to the gradient update to determine which parameters are to be updated while we introduce dense learnable tensors to model second-order dependencies with a series of tensor products.


\textbf{Few-shot classification:} As a good test bed to evaluate few-shot learning, huge progress has been made in the few-shot classification task. Triggered by \cite{matchingnet}, many recent studies have focused on discovering effective inductive bias on classification task. For example, network architectures that perform nearest neighbor search \cite{matchingnet, Snell-nips-2017} were suggested. Some improved the performance by modeling the interactions or correlation between training examples \cite{Mishra-iclr18, Garcia-iclr18, Sung-cvpr18, Oreshkin-nips18, Munkhdalai-icml18}. In order to overcome the nature of few-shot learning, the generative models have been suggested to augment the training data \cite{Schwartz-nips18,Wang-cvpr18} or generate model parameters for the specified task \cite{rusu-iclr19,Qiao-cvpr18}. The state-of-the-art results are achieved by additionally training 64-way classification task for pretraining \cite{Qiao-cvpr18, rusu-iclr19, Oreshkin-nips18} with larger ResNet models \cite{Qiao-cvpr18, rusu-iclr19, Munkhdalai-icml18, Mishra-iclr18}. In this work, our focus is to improve the model-agnostic few-shot learner that is broadly applicable to other tasks, e.g. reinforcement learning setup.

\textbf{Learning optimizers:}
Our proposed method may fall within the \textit{learning optimizer} category \cite{Ravi-iclr-2017, Andrychowicz-nips-2016,Wichrowska-icml17,Metz-arxiv-2018}. They also take as input the gradient and transform it via a neural network to achieve better convergence behavior. However, their main focus is to capture the training dynamics of individual gradient coordinates \cite{Ravi-iclr-2017, Andrychowicz-nips-2016} or to obtain a generic optimizer that is broadly applicable for different datasets and architectures \cite{Wichrowska-icml17,Metz-arxiv-2018,Andrychowicz-nips-2016}. On the other hand, we meta-learn a curvature coupled with the model’s initialization parameters. We focus on a fast adaptation scenario requiring a small number of gradient steps. Therefore, our method does not consider a history of the gradients, which enables us to avoid considering a complex recurrent architecture. Finally, our approach is well connected to existing second order methods while learned optimizers are not easily interpretable since the gradient passes through nonlinear and multilayer recurrent neural networks.

\section{Experiments}
\label{experiments}
We evaluate the proposed method on a synthetic data few-shot regression task few-shot image classification tasks with Omniglot and MiniImagenet datasets. We test two versions of the meta-curvature. The first one, named as MC1, we fixed the $\B{M}_o = \B{I}$ Eq. \ref{eq:mc_tensor_expand}. The second one, named as MC2, we learn all three meta-curvature matrices. We also report results on few-shot reinforcement learning in appendices.

\subsection{Few-shot regression}
\begin{wraptable}{r}{6.5cm}
\caption{Few-shot regression results.}
\label{table:regression}
\begin{small}
\begin{tabular}{lcc}
\toprule
Method & 5-shot & 10-shot \\
\midrule
MAML               & 0.686 $\pm$ 0.070 & 0.435 $\pm$ 0.039 \\
Meta-SGD           & 0.482 $\pm$ 0.061 & 0.258 $\pm$ 0.026 \\
LayerLR            & 0.528 $\pm$ 0.068 & 0.269 $\pm$ 0.027 \\
MC1                & 0.426 $\pm$ 0.054 & 0.239 $\pm$ 0.025 \\
MC2                & \textbf{0.405} $\pm$ \textbf{0.048} & \textbf{0.201} $\pm$ \textbf{0.020} \\
\bottomrule
\end{tabular}
\end{small}
\end{wraptable}

To begin with, we perform a simple regression problem following \cite{Finn-icml-2017,Li-arxiv-2017}. During the meta-training process, sinusoidal functions are sampled, where the amplitude and phase are varied within $[0.1,5.0]$ and $[0,\pi]$ respectively. The network architecture and all hyperparameters are same as \cite{Finn-icml-2017} and we only introduce the suggested meta-curvature. We reported the mean squared error with 95\% confidence interval after one gradient step in Figure \ref{table:regression}. The details are provided in appendices.

\subsection{Few-shot classification on Omniglot}
\begin{table}[t]
\caption{Few-shot classification results on Omniglot dataset. $^\dagger$ denotes 3 model ensemble. }
\label{table:omniglot}
\begin{center}
\begin{footnotesize}
\begin{tabular}{lcccc}
\toprule
& 5-way 1-shot & 5-way 5-shot & 20-way 1-shot & 20-way 5-shot \\
\midrule
SNAIL \cite{snail} & 99.07 $\pm$ 0.16 & 99.78 $\pm$ 0.09 & 97.64 $\pm$ 0.30 & 99.36 $\pm$ 0.18 \\
GNN \cite{gnn} & 99.2 & 99.7 & 97.4 & 99.0 \\
\midrule
MAML & 98.7 $\pm$ 0.4 & 99.9 $\pm$ 0.1 & 95.8 $\pm$ 0.3 & 98.9 $\pm$ 0.2 \\
Meta-SGD & 99.53 $\pm$ 0.26 & 99.93 $\pm$ 0.09 & 95.93 $\pm$ 0.38 & 98.97 $\pm$ 0.19 \\
MAML++$^\dagger$ \cite{Antoniou-arxiv-2018} & 99.47  & 99.93 & 97.65 $\pm$ 0.05 & 99.33 $\pm$ 0.03 \\
MC1 & 99.47 $\pm$ 0.27 & 99.57 $\pm$ 0.12 & 97.60 $\pm$ 0.29 & 99.23 $\pm$ 0.08 \\
MC2 & 99.77 $\pm$ 0.17 & 99.79 $\pm$ 0.10 & 97.86 $\pm$ 0.26 & 99.24 $\pm$ 0.07 \\
MC2$^\dagger$ & \textbf{99.97} $\pm$ \textbf{0.06} & 99.89 $\pm$ 0.06 & \textbf{99.12} $\pm$ \textbf{0.16} & \textbf{99.65} $\pm$ \textbf{0.05} \\
\bottomrule
\end{tabular}
\end{footnotesize}
\end{center}
\vskip -0.1in
\end{table}

\begin{figure*}[t]
\begin{center}
\centerline{\includegraphics[width=\linewidth]{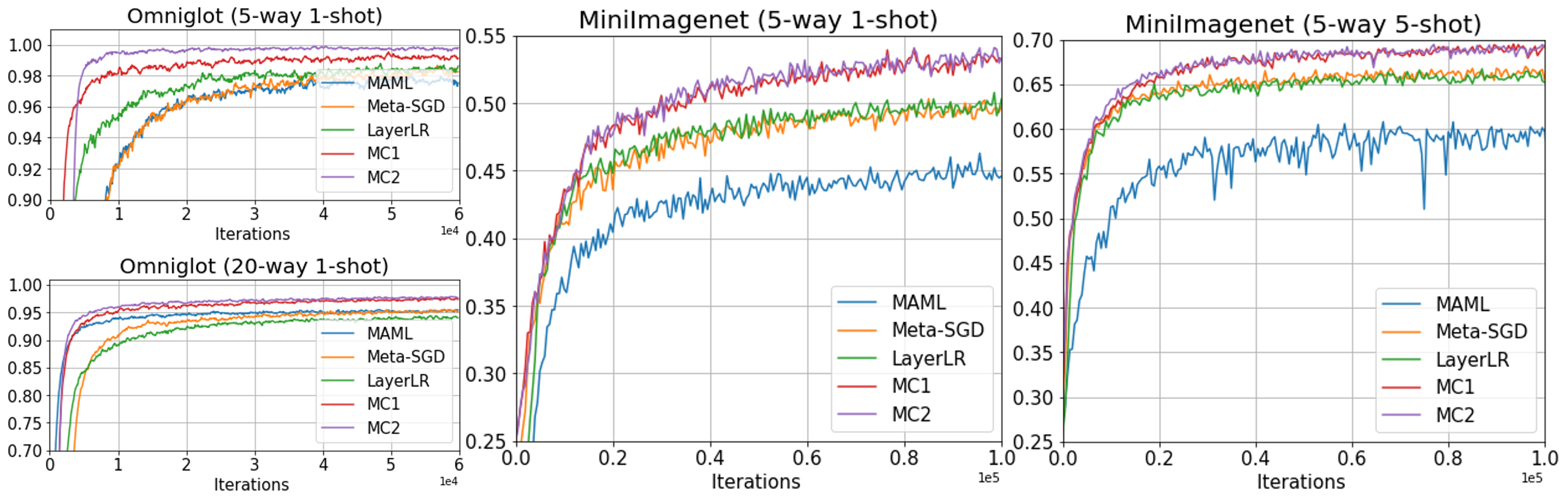}}
\caption{Few-shot classification accuracy over training iterations.}
\label{fig:cls_experiment}
\end{center}
\end{figure*}

The Omniglot dataset consists of handwritten characters from 50 different languages and 1632 different characters. It has been widely used to evaluate few-shot classification performance. We follow the experimental protocol in \cite{Finn-icml-2017} and all hyperparameters and network architecture are same as \cite{Finn-icml-2017}. Further experimental details are provided in appendices. Except 5-shot 5-way setting, our simple 4 layers CNN with meta-curvatures outperform all MAML variants and also achieved state-of-the-art results without additional specialized architectures, such as attention module (SNAIL \cite{snail}) or relational module (GNN \cite{gnn}). We provide the training curves in Figure \ref{fig:cls_experiment} and our methods converge much faster and achieve higher accuracy.

\begin{table}[t]
\caption{Few-shot classification results on miniImagenet test set (5-way classification) with baseline 4 layer CNNs. \mbox{*} is from the original papers. $^\dagger$ denotes 3 model ensembles.}
\vskip -0.1in
\label{table:miniimagenet}
\begin{center}
\begin{small}
\begin{tabular}{lcccc}
\toprule
& \multicolumn{2}{c}{1-shot} & \multicolumn{2}{c}{5-shot} \\
\midrule
Inner steps & 1 step & 5 step & 1 step & 5 step \\
\midrule
\mbox{*}MAML               & $\cdot$  & 48.7 $\pm$ 1.84    & $\cdot$   & 63.1 $\pm$ 0.92\\
\mbox{*}Meta-SGD           & 50.47 $\pm$ 1.87   & $\cdot$           & 64.03 $\pm$ 0.94 & $\cdot$  \\
\mbox{*}MAML++$^\dagger$  & 51.05 $\pm$ 0.31   & 52.15 $\pm$ 0.26   & $\cdot$           & 68.32 $\pm$ 0.44 \\
\midrule
MAML       & 46.28 $\pm$ 0.89   & 48.85 $\pm$ 0.88    & 59.26 $\pm$ 0.72  & 63.92 $\pm$ 0.74 \\
Meta-SGD   & 49.87 $\pm$ 0.87   & 48.99 $\pm$ 0.86    & 66.35 $\pm$ 0.72  & 63.84 $\pm$ 0.71 \\
LayerLR            & 50.04 $\pm$ 0.87   & 50.55 $\pm$ 0.87    & 65.06 $\pm$ 0.71  & 66.64 $\pm$ 0.69 \\
MC1                & 53.37 $\pm$ 0.88   & 53.74 $\pm$ 0.84    & \textbf{68.47} $\pm$ \textbf{0.69} & \textbf{68.01} $\pm$ \textbf{0.73} \\
MC2                & \textbf{54.23} $\pm$ \textbf{0.88}   & \textbf{54.08} $\pm$ \textbf{0.93} & 67.94 $\pm$ 0.71  & \textbf{67.99} $\pm$ \textbf{0.73} \\
MC2$^\dagger$     & \textbf{54.90} $\pm$ \textbf{0.90}   & \textbf{55.73} $\pm$ \textbf{0.94} & \textbf{69.46} $\pm$ \textbf{0.70}  & \textbf{70.33} $\pm$ \textbf{0.72} \\
\bottomrule
\end{tabular}
\end{small}
\end{center}
\vskip -0.1in
\end{table}

\begin{table}[h]
\vskip -0.1in
\caption{The results on miniImagenet and tieredImagenet. $^\ddagger$ indicates that both meta-train and meta-validation are used during meta-training. $^\dagger$ denotes indicates that 15-shot meta-training was used for both 1-shot and 5-shot testing. MetaOptNet [3] used ResNet-12 backbone and trained end-to-end manner while we used the fixed features provided by [2] (center - features from the central crop, multiview - features averaged over four corners, central crops, and horizontal mirrored).}
\vskip -0.1in
\label{table:miniimagenet_wrn}
\begin{center}
\begin{small}
\begin{tabular}{lccccc}
\toprule
& \multicolumn{2}{c}{miniImagenet} & \multicolumn{2}{c}{tieredImagenet} \\
\midrule
& 1-shot & 5-shot & 1-shot & 5-shot \\
\midrule
\cite{Qiao-cvpr18}$^\ddagger$ & 59.60 $\pm$ 0.41 & 73.74 $\pm$ 0.19 & $\cdot$ & $\cdot$ \\
LEO (center)$^\ddagger$ \cite{rusu-iclr19} & 61.76 $\pm$ 0.08 & 77.59 $\pm$ 0.12 & 66.33 $\pm$ 0.05 & 81.44 $\pm$ 0.09 \\
LEO (multiview)$^\ddagger$ \cite{rusu-iclr19} & 63.97 $\pm$ 0.20 & 79.49 $\pm$ 0.70 & $\cdot$ & $\cdot$ \\
MetaOptNet-SVM$^\ddagger$$^\dagger$ \cite{metaopt} & 64.09 $\pm$ 0.62 & 80.00 $\pm$ 0.45 & 65.81 $\pm$ 0.74 & 81.75 $\pm$ 0.53 \\
\midrule
Meta-SGD (center)  & 56.58 $\pm$ 0.21 &  68.84 $\pm$ 0.19 & 59.75 $\pm$ 0.25 & 69.04 $\pm$ 0.22 \\
MC2 (center)       & 61.22 $\pm$ 0.10 &  75.92 $\pm$ 0.17 & 66.20 $\pm$ 0.10 & 82.21 $\pm$ 0.08 \\
MC2 (center)$^\ddagger$ & \textbf{61.85} $\pm$ \textbf{0.10} &  77.02 $\pm$ 0.11 & \textbf{67.21} $\pm$ \textbf{0.10} & \textbf{82.61} $\pm$ \textbf{0.08} \\
MC2 (multiview)$^\ddagger$ &  \textbf{64.40} $\pm$ \textbf{0.10}  & \textbf{80.21} $\pm$ \textbf{0.10} & $\cdot$ & $\cdot$ \\
\bottomrule
\end{tabular}
\end{small}
\end{center}
\vskip -0.1in
\end{table}

\subsection{Few-shot classification on miniImagenet and tieredImagenet}

\textbf{Datasets:} The miniImagenet dataset was proposed by \cite{matchingnet,Ravi-iclr-2017} and it consists of 100 subclasses out of 1000 classes in the original dataset (64 training classes, 12 validation classes, 24 test classes). The tieredImagenet dataset \cite{ren-iclr18} is a larger subset, composed of 608 classes and reduce the semantic similarity between train/val/test splits by considering high-level categories.

\textbf{baseline CNNs:} We used 4 layers convolutional neural network with the batch normalization followed by a fully connected layer for the final classification. In order to increase the capacity of the network, we increased the filter size up to 128. We found that the model with the larger filter seriously overfit (also reported in \cite{Finn-icml-2017}). To avoid overfitting, we applied data augmentation techniques suggested in \cite{autoaugment, cutout}. For a fair comparison to \cite{Antoniou-arxiv-2018}, we also reported the results of model ensemble. Throughout the meta-training, we saved the model regularly and picked 3 models that have the best accuracy on the meta-validation dataset. We re-implemented all three baselines and performed the experiments with the same settings. We provide further the details in the appendices.

Fig. \ref{fig:cls_experiment} and Table \ref{table:miniimagenet} shows the results of baseline CNNs experiments on miniImagenet. MC1 and MC2 outperformed all other baselines for all different experiment settings. Not only does MC reach a higher accuracy at convergence, but also showed a much faster convergence rates for meta-training. Our methods share the same benefits as second order methods although we do not approximate any Hessian or Fisher matrices. Unlike other MAML variants, which required an extensive hyperparameter search, our methods are very robust to hyperparameter settings. Usually, MC2 outperforms MC1 because the more fine-grained meta-curvature enable us to effectively increase the model's capacity.

\textbf{WRN-28-10 features and MLP:} To the best of our knowledge, \cite{rusu-iclr19, Qiao-cvpr18} are current state-of-the-art methods that use a pretrained WRN-28-10 \cite{wrn} network (trained with 64-way classification task on entire meta-training set) as a feature extractor network. We evaluated our methods on this setting by adding one hidden layer MLP followed by a softmax classifier and our method again improved MAML variants by a large margin. Despite our best attempts, we could not find a good hyperparameters to train original MAML in this setting. 
Although our main goal is to push how much a simple gradient transformation in the inner loop optimization can improve general and broadly applicable MAML frameworks, our methods outperformed the recent methods that used various task specific techniques, e.g. task dependent weight generating methods \cite{rusu-iclr19, Qiao-cvpr18} and relational networks \cite{rusu-iclr19}. Our methods also outperformed the very latest state of the art results \cite{metaopt} that used extensive data-augmentation, regularization, and 15-shot meta-training schemes with different backbone networks.



\section{Conclusion}
\label{conclusion}
We propose to meta-learn the curvature for faster adaptation and better generalization. The suggested method significantly improved the performance upon previous MAML variants and outperformed the recent state of the art methods. It also leads to faster convergence during meta-training. We present an analysis about generalization performance and connect to existing second order methods, which would provide useful insights for further research.



\bibliography{meta_curvature.bib}

\begin{thebibliography}{10}

\bibitem{Shedivat-iclr18}
Maruan Al-Shedivat, Trapit Bansal, Yuri Burda, Ilya Sutskever, Igor Mordatch,
  and Pieter Abbeel.
\newblock {Continuous Adaptation via Meta-Learning in Nonstationary and
  Competitive Environments}.
\newblock In {\em International Conference on Learning Representations (ICLR)},
  2018.

\bibitem{Amari}
Shun-Ichi Amari.
\newblock {Natural gradient works efficiently in learning}.
\newblock {\em Neural computation}, 10(2):251--276, 1998.

\bibitem{Andrychowicz-nips-2016}
Marcin Andrychowicz, Misha Denil, Sergio~Gómez Colmenarejo, Matthew~W.
  Hoffman, David Pfau, Tom Schaul, Brendan Shillingford, and Nando~de Freitas.
\newblock Learning to learn by gradient descent by gradient descent.
\newblock In {\em Neural Information Processing Systems (NeurIPS)}, 2016.

\bibitem{Antoniou-arxiv-2018}
Antreas Antoniou, Harrison Edwards, and Amos Storkey.
\newblock {How to train your MAML}.
\newblock {\em International Conference on Learning Representations (ICLR)},
  2019.

\bibitem{autoaugment}
Ekin~D. Cubuk, Barret Zoph, Dandelion Mane, Vijay Vasudevan, and Quoc~V. Le.
\newblock {AutoAugment: Learning Augmentation Policies from Data}.
\newblock {\em arXiv:1805.09501}, 2018.

\bibitem{cutout}
Terrance DeVries and Graham~W. Taylor.
\newblock {Improved regularization of convolutional neural networks with
  cutout}.
\newblock {\em arXiv:1708.04552}, 2017.

\bibitem{Du-arxiv-2018}
Yunshu Du, Wojciech~M. Czarnecki, Siddhant~M. Jayakumar, Razvan Pascanu, and
  Balaji Lakshminarayanan.
\newblock {Adapting Auxiliary Losses Using Gradient Similarity}.
\newblock {\em arXiv:1812.02224}, 2018.

\bibitem{adagrad}
John Duchi, Elad Hazan, and Yoram Singer.
\newblock Adaptive subgradient methods for online learning and stochastic
  optimization.
\newblock {\em Journal of Machine Learning Research}, 12:2121--2159, 2011.

\bibitem{Finn-icml-2017}
Chelsea Finn, Pieter Abbeel, and Sergey Levine.
\newblock {Model-Agnostic Meta-Learning for Fast Adaptation of Deep Networks}.
\newblock In {\em International Conference on Machine Learning (ICML)}, 2017.

\bibitem{Finn-iclr-2018}
Chelsea Finn and Sergey Levine.
\newblock {Meta-Learning and Universality: Deep Representations and Gradient
  Descent Can Approximate Any Learning Algorithm}.
\newblock In {\em International Conference on Learning Representations (ICLR)},
  2018.

\bibitem{Garcia-iclr18}
Victor Garcia and Joan Bruna.
\newblock {Few-Shot Learning with Graph Neural Networks}.
\newblock In {\em International Conference on Learning Representations (ICLR)},
  2018.

\bibitem{gnn}
Victor Garcia and Joan Bruna.
\newblock {Few-Shot Learning with Graph Neural Networks}.
\newblock In {\em International Conference on Learning Representations (ICLR)},
  2018.

\bibitem{Grant-iclr-2018}
Erin Grant, Chelsea Finn, Sergey Levine, Trevor Darrell, and Thomas Griffiths.
\newblock {Recasting Gradient-Based Meta-Learning as Hierarchical Bayes}.
\newblock In {\em International Conference on Learning Representations (ICLR)},
  2018.

\bibitem{Grosse-icml-2016}
Roger Grosse and James Martens.
\newblock {A Kronecker-factored approximate Fisher matrix for convolution
  layers}.
\newblock In {\em International Conference on Machine Learning (ICML)}, 2016.

\bibitem{Kim-nips2018}
Taesup Kim, Jaesik Yoon, Ousmane Dia, Sungwoong Kim, Yoshua Bengio, and Sungjin
  Ahn.
\newblock {Bayesian Model-Agnostic Meta-Learning}.
\newblock In {\em Neural Information Processing Systems (NeurIPS)}, 2018.

\bibitem{adam}
Diederik~P. Kingma and Jimmy~Lei Ba.
\newblock Adam: A method for stochastic optimization.
\newblock In {\em International Conference on Learning Representations (ICLR)},
  2015.

\bibitem{KoBa09}
Tamara~G. Kolda and Brett~W. Bader.
\newblock {Tensor Decompositions and Applications}.
\newblock {\em SIAM Review}, 51(3):455--500, 2009.

\bibitem{Kossaif-arxiv-2018}
Jean Kossaif, Zachary Lipton, Aran Khanna, Tommaso Furlanello, and Anima
  Anandkumar.
\newblock {Tensor Regression Networks}.
\newblock {\em arXiv:1707.08308}, 2018.

\bibitem{Lake1332}
Brenden~M. Lake, Ruslan Salakhutdinov, and Joshua~B. Tenenbaum.
\newblock Human-level concept learning through probabilistic program induction.
\newblock {\em Science}, 350(6266):1332--1338, 2015.

\bibitem{metaopt}
Kwonjoon Lee, Subhransu Maji, Avinash Ravichandran, and Stefano Soatto.
\newblock {Meta-Learning with Differentiable Convex Optimization}.
\newblock In {\em IEEE Conference on Computer Vision and Pattern Recognition
  (CVPR)}, 2019.

\bibitem{lee-icml-2018}
Yoonho Lee and Seungjin Choi.
\newblock Gradient-based meta-learning with learned layerwise metric and
  subspace.
\newblock In {\em International Conference on Machine Learning (ICML)}, 2018.

\bibitem{Li-arxiv-2017}
Zhenguo Li, Fengwei Zhou, Fei Chen, and Hang Li.
\newblock {Meta-SGD: Learning to Learn Quickly for Few Shot Learning}.
\newblock {\em arXiv:1707.09835}, 2017.

\bibitem{Lopez-nips2017}
David Lopez-Paz and Marc'Aurelio Ranzato.
\newblock {Gradient Episodic Memory for Continual Learning}.
\newblock In {\em Neural Information Processing Systems (NeurIPS)}, 2017.

\bibitem{Martens-icml-2015}
James Martens and Roger Grosse.
\newblock {Optimizing Neural Networks with Kronecker-factored Approximate
  Curvature}.
\newblock In {\em International Conference on Machine Learning (ICML)}, 2015.

\bibitem{Metz-arxiv-2018}
Luke Metz, Niru Maheswaranathan, Jeremy Nixon, C.~Daniel Freeman, and Jascha
  Sohl-Dickstein.
\newblock {Learned Optimizers That Outperform SGD On Wall-Clock And Test Loss}.
\newblock {\em arXiv:1810.10180}, 2018.

\bibitem{Mishra-iclr18}
Nikhil Mishra, Mostafa Rohaninejad, Xi~Chen, and Pieter Abbeel.
\newblock {A Simple Neural Attentive Meta-Learner}.
\newblock In {\em International Conference on Learning Representations (ICLR)},
  2018.

\bibitem{snail}
Nikhil Mishra, Mostafa Rohaninejad, Xi~Chen, and Pieter Abbeel.
\newblock {A Simple Neural Attentive Meta-Learner}.
\newblock In {\em International Conference on Learning Representations (ICLR)},
  2018.

\bibitem{Muandet-nips2012}
Krikamol Muandet, Kenji Fukumizu, Francesco Dinuzzo, and Bernhard Schölkopf.
\newblock {Learning from Distributions via Support Measure Machines}.
\newblock In {\em Neural Information Processing Systems (NeurIPS)}, 2012.

\bibitem{Munkhdalai-icml18}
Tsendsuren Munkhdalai, Xingdi Yuan, Soroush Mehri, and Adam Trischler.
\newblock {Rapid Adaptation with Conditionally Shifted Neurons}.
\newblock In {\em International Conference on Machine Learning (ICML)}, 2018.

\bibitem{Nichol-arxiv-2018}
Alex Nichol, Joshua Achiam, and John Schulman.
\newblock {On First-Order Meta-Learning Algorithms}.
\newblock {\em arXiv:1803.02999}, 2018.

\bibitem{Nocedal_2006}
Jorge Nocedal and Stephen Wright.
\newblock {\em {Numerical Optimization}}.
\newblock Springer Science \& Business Media, 2006.

\bibitem{Oreshkin-nips18}
Boris~N. Oreshkin, Pau Rodriguez, and Alexandre Lacoste.
\newblock {TADAM: Task dependent adaptive metric for improved few-shot
  learning}.
\newblock In {\em Neural Information Processing Systems (NeurIPS)}, 2018.

\bibitem{Qiao-cvpr18}
Siyuan Qiao, Chenxi Liu, Wei Shen, and Alan Yuille.
\newblock {Few-Shot Image Recognition by Predicting Parameters from
  Activations}.
\newblock In {\em IEEE Conference on Computer Vision and Pattern Recognition
  (CVPR)}, 2018.

\bibitem{Ravi-iclr-2017}
Sachin Ravi and Hugo Larochelle.
\newblock {Optimization As a Model For Few-shot Learning}.
\newblock In {\em International Conference on Learning Representations (ICLR)},
  2017.

\bibitem{ren-iclr18}
Mengye Ren, Eleni Triantafillou, Sachin Ravi, Jake Snell, Kevin Swersky,
  Joshua~B. Tenenbaum, Hugo Larochelle, and Richard~S. Zemel.
\newblock {Meta-Learning for Semi-Supervised Few-Shot Classification}.
\newblock In {\em International Conference on Learning Representations (ICLR)},
  2018.

\bibitem{Riemer-iclr19}
Matthew Riemer, Ignacio Cases, Robert Ajemian, Miao Liu, Irina Rish, Yuhai Tu,
  and Gerald Tesauro.
\newblock {Learning to Learn without Forgetting By Maximizing Transfer and
  Minimizing Interference}.
\newblock In {\em International Conference on Learning Representations (ICLR)},
  2019.

\bibitem{promp}
Jonas Rothfuss, Dennis Lee, Ignasi Clavera, Tamim Asfour, and Pieter Abbeel.
\newblock Promp: Proximal meta-policy search.
\newblock In {\em International Conference on Learning Representations (ICLR)},
  2019.

\bibitem{LeRoux-2008}
Nicolas~Le Roux, Pierre-Antoine Manzagol, and Yoshua Bengio.
\newblock {Topmoumoute online natural gradient algorithm}.
\newblock In {\em Neural Information Processing Systems (NeurIPS)}, 2008.

\bibitem{rusu-iclr19}
Andrei~A. Rusu, Dushyant Rao, Jakub Sygnowski, Oriol Vinyals, Razvan Pascanu,
  Simon Osindero, and Raia Hadsell.
\newblock {Meta-Learning with Latent Embedding Optimization}.
\newblock In {\em International Conference on Learning Representations (ICLR)},
  2019.

\bibitem{trpo}
John Schulman, Sergey Levine, Philipp Moritz, and Pieter~Abbeel Michael
  I.~Jordan.
\newblock Trust region policy optimization.
\newblock In {\em International Conference on Machine Learning (ICML)}, 2015.

\bibitem{ppo}
John Schulman, Filip Wolski, Prafulla Dhariwal, Alec Radford, and Oleg Klimov.
\newblock {Proximal Policy Optimization Algorithms}.
\newblock {\em arXiv:1707.06347}, 2017.

\bibitem{Schwartz-nips18}
Eli Schwartz, Leonid Karlinsky, Joseph Shtok, Sivan Harary, Mattias Marder,
  Rogerio Feris, Abhishek Kumar, Raja Giryes, and Alex~M. Bronstein.
\newblock {Delta-encoder: an effective sample synthesis method for few-shot
  object recognition}.
\newblock In {\em Neural Information Processing Systems (NeurIPS)}, 2018.

\bibitem{Snell-nips-2017}
Jake Snell, Kevin Swersky, and Richard~S. Zemel.
\newblock {Prototypical Networks for Few-shot Learning}.
\newblock In {\em Neural Information Processing Systems (NeurIPS)}, 2017.

\bibitem{Sung-cvpr18}
Flood Sung, Yongxin Yang, Li~Zhang, Tao Xiang, and Timothy M.~Hospedales Philip
  H.S.~Torr.
\newblock {Learning to Compare: Relation Network for Few-Shot Learning}.
\newblock In {\em IEEE Conference on Computer Vision and Pattern Recognition
  (CVPR)}, 2018.

\bibitem{rmsprop}
Tijmen Tieleman and Geoffrey Hinton.
\newblock {Lecture 6.5---RmsProp: Divide the gradient by a running average of
  its recent magnitude}.
\newblock COURSERA: Neural Networks for Machine Learning, 2012.

\bibitem{mujoco}
Emanuel Todorov, Tom Erez, and Yuval Tassa.
\newblock Mujoco: A physics engine for model-based control.
\newblock In {\em International Conference on Intelligent Robots and Systems
  (IROS)}, 2012.

\bibitem{matchingnet}
Oriol Vinyals, Charles Blundell, Timothy Lillicrap, Koray Kavukcuoglu, and Daan
  Wierstra.
\newblock {Matching Networks for One Shot Learning}.
\newblock In {\em Neural Information Processing Systems (NeurIPS)}, 2016.

\bibitem{Wang-cvpr18}
Yu-Xiong Wang, Ross Girshick, Martial Hebert, and Bharath Hariharan.
\newblock {Low-Shot Learning from Imaginary Data}.
\newblock In {\em IEEE Conference on Computer Vision and Pattern Recognition
  (CVPR)}, 2018.

\bibitem{Wichrowska-icml17}
Olga Wichrowska, Niru Maheswaranathan, Matthew~W. Hoffman, Sergio~Gomez
  Colmenarejo, Misha Denil, Nando de~Freitas, and Jascha Sohl-Dickstein.
\newblock {Learned Optimizers that Scale and Generalize}.
\newblock In {\em International Conference on Machine Learning (ICML)}, 2017.

\bibitem{wrn}
Sergey Zagoruyko and Nikos Komodakis.
\newblock {Wide Residual Networks}.
\newblock In {\em The British Machine Vision Conference (BMVC)}, 2016.

\end{thebibliography}
\bibliographystyle{plain}

\clearpage
\appendix

\section{Meta-training algorithm}
Alg. \ref{alg:meta_curvature_alg} shows the details of the algorithm to train meta-curvature matrices and the initial model parameters. To avoid cluttered notation, we assumed the model has only one layer and it is straightforward to extend to multiple layers.

\begin{algorithm}
\footnotesize
   \caption{Training MAML with the meta-curvature for few-shot supervised learning}
   \label{alg:meta_curvature_alg}
\begin{algorithmic}
\footnotesize
\STATE {Input:} task distribution $p(\mathcal{T})$, learning rate $\alpha$, $\beta$
\STATE Initialize $ \mathbf{M}_o, \mathbf{M}_i, \mathbf{M}_f = \mathbf{I}$
\WHILE{not converged}
    \STATE Sample batch of tasks $\tau_i \sim p(\mathcal{T})$
    \FOR{\textbf{all}  $\tau_i$ \textbf{do}}
        \STATE $\theta^{\tau_i} = \theta - \alpha \mathbf{M}_{mc} \nabla \mathcal{L}_{\textrm{tr}}^{\tau_i}(\theta)$ \COMMENT{Assuming one gradient step}
    \ENDFOR
    \STATE $\theta \leftarrow \textrm{ADAM} \big( \theta, \beta, \nabla_\theta \sum_{\tau_i} \mathcal{L}_{\textrm{val}}^{\tau_i}(\theta^{\tau_i}) \big)$
    \STATE $\mathbf{M}_o \leftarrow \textrm{ADAM} \big( \mathbf{M}_o, \beta, \nabla_{\mathbf{M}_o} \sum_{\tau_i} \mathcal{L}_{\textrm{val}}^{\tau_i}(\theta^{\tau_i}) \big)$
    \STATE $\mathbf{M}_i \leftarrow \textrm{ADAM} \big( \mathbf{M}_i, \beta, \nabla_{\mathbf{M}_i} \sum_{\tau_i} \mathcal{L}_{\textrm{val}}^{\tau_i}(\theta^{\tau_i}) \big)$
    \STATE $\mathbf{M}_f \leftarrow \textrm{ADAM} \big( \mathbf{M}_f, \beta, \nabla_{\mathbf{M}_f} \sum_{\tau_i} \mathcal{L}_{\textrm{val}}^{\tau_i}(\theta^{\tau_i}) \big)$
\ENDWHILE
\end{algorithmic}
\end{algorithm}

\section{Few-shot regression}

\begin{table}[t]
\caption{Few-shot regression results on sinusoidal functions.}
\label{table:regression}
\begin{center}
\begin{small}
\begin{tabular}{lccc}
\toprule
Method & 5-shot & 10-shot & 20-shot \\
\midrule
MAML               & 0.686 $\pm$ 0.070 & 0.435 $\pm$ 0.039 & 0.228 $\pm$ 0.024 \\
Meta-SGD           & 0.482 $\pm$ 0.061 & 0.258 $\pm$ 0.026 & 0.127 $\pm$ 0.013 \\
LayerLR            & 0.528 $\pm$ 0.068 & 0.269 $\pm$ 0.027 & 0.134 $\pm$ 0.014 \\
MC1                & 0.426 $\pm$ 0.054 & 0.239 $\pm$ 0.025 & 0.125 $\pm$ 0.013 \\
MC2                & \textbf{0.405} $\pm$ \textbf{0.048} & \textbf{0.201} $\pm$ \textbf{0.020} & \textbf{0.112} $\pm$ \textbf{0.011} \\
\bottomrule
\end{tabular}
\end{small}
\end{center}
\end{table}

\textbf{Experimental setup}
We used the same experimental setups in \cite{Finn-icml-2017}. During training and testing, the amplitude and the phase vary within $[0.1, 5.0]$ and $[0,\pi]$ respectively, and data points are sampled from uniform distribution $[-5,5]$. We used one gradient step with the fixed learning rate 0.01 and Adam was used for meta-training with the outer loop learning rate 0.001. We used the same network architecture, which has two 40 dimension fully connected layers with ReLU activation. We sampled 25 tasks for every iterations and trained 70000 iterations. We reported the performance from the trained model that had the minimum loss value. \cite{Finn-icml-2017} reported the MSE for 5-shot setting, and we could reproduced the results. \cite{Li-arxiv-2017} has slightly different settings, so the MSE are not directly comparable to theirs.

\begin{figure}[t]
\begin{center}
\centerline{\includegraphics[height=20cm]{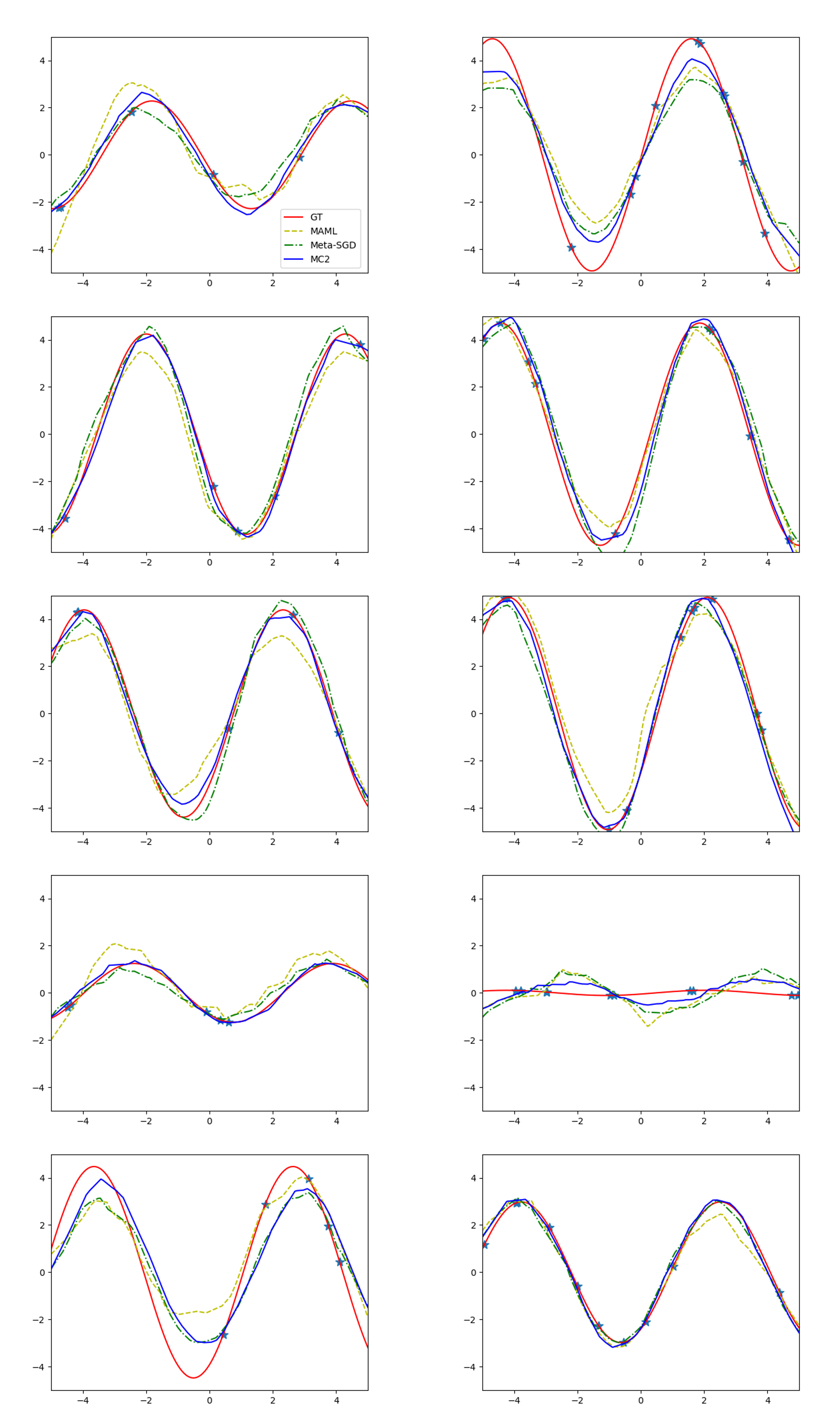}}
\caption{Qualitative results of few-shot regression on sinusoidal functions. The left column - 5 shot, The right column - 10 shot}
\label{fig:sinosoid}
\end{center}
\end{figure}

\textbf{Qualitative results}: We provide qualitative results of few-shot regression task on sinusoidal functions in Figure \ref{fig:sinosoid}. The star shape markers are the few data points for training, and we draw the curves based on each methods, MAML, Meta-SGD, and the proposed MC2. The left column is 5-shot and the right column is 10-shot experiments.

\begin{figure*}[t]
\begin{center}
\centerline{\includegraphics[width=0.7\linewidth]{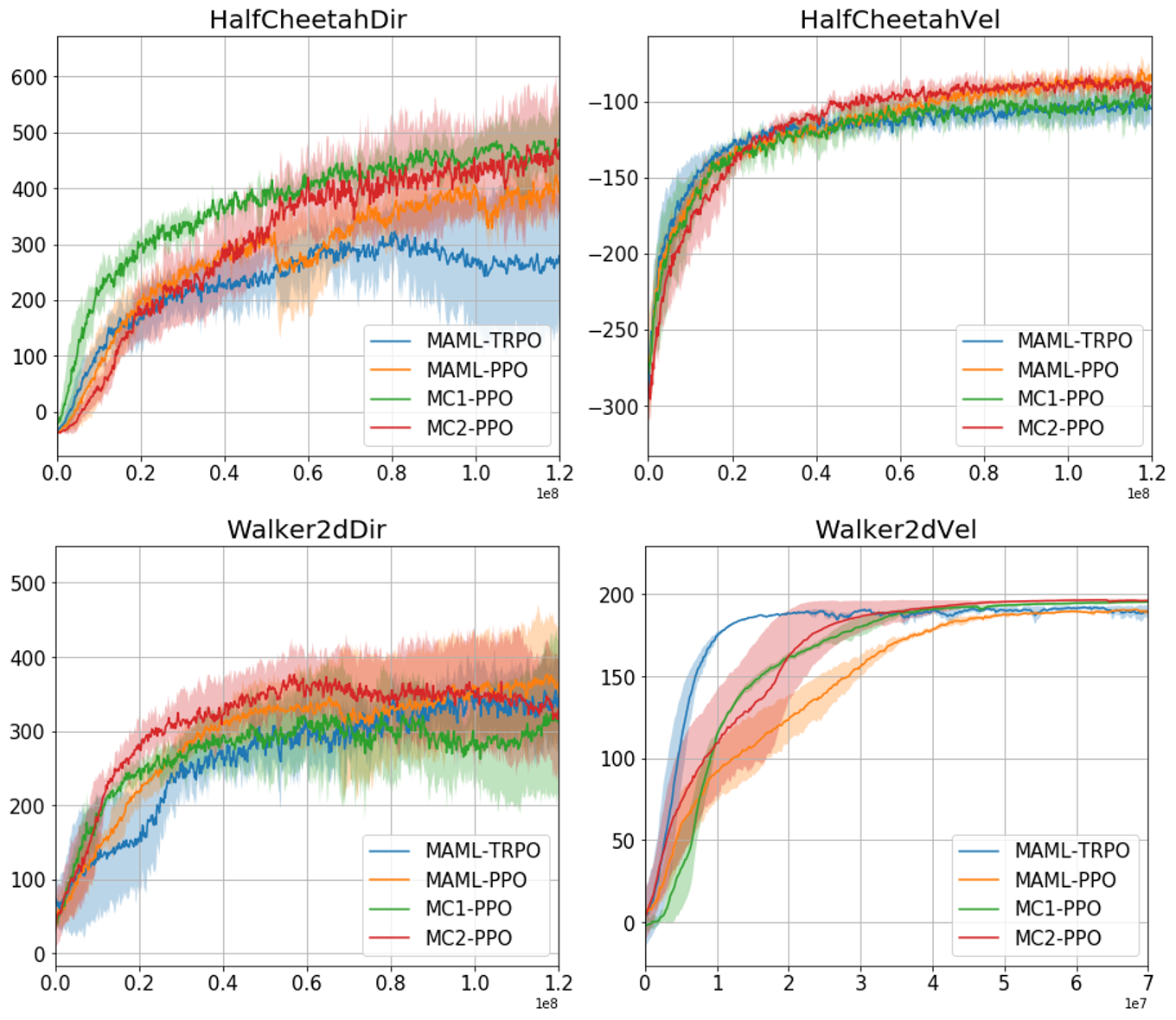}}
\caption{Reinforcement learning experimental results. Y-axis: rewards after the model updates. X-axis: meta-training steps. We performed at least three runs with random seeds and the curves are averaged over them.}
\label{fig:rl_experiment}
\end{center}
\end{figure*}

\begin{figure}[t]
\begin{center}
\centerline{\includegraphics[width=0.8\linewidth]{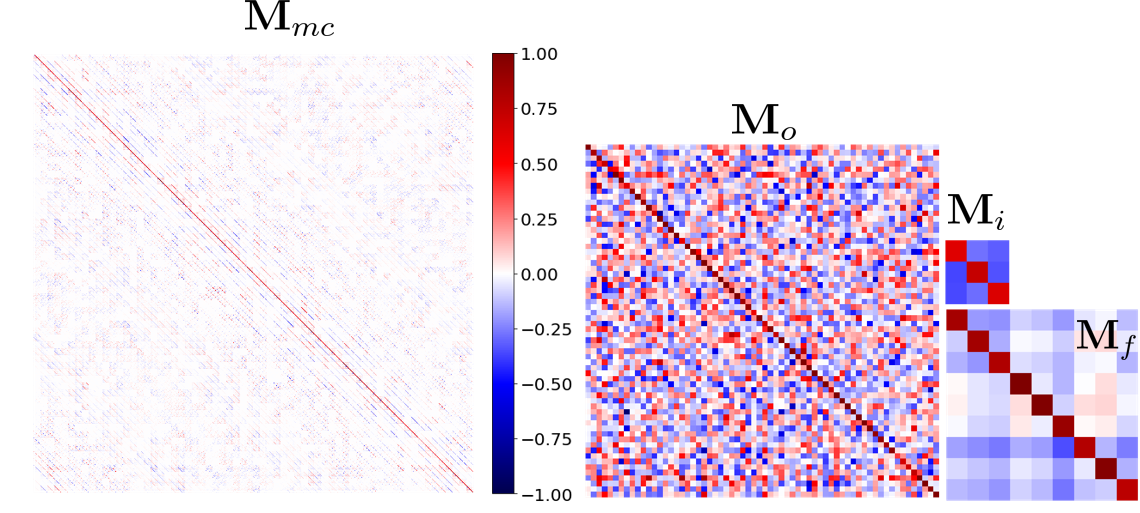}}
\caption{Visualization of meta-curvature matrices. We clipped the values $[-1,1]$ for better visualization (Best viewed in color)}
\label{fig:meta_curvature_vis}
\end{center}
\end{figure}

\section{Few-shot classification on Omniglot dataset}
We used the same experimental setups in \cite{Finn-icml-2017}. Out of 1623 characters, we used 1100 characters for training, 100 characters for validation, and remaining 423 characters for testing. The network architecture is 4 convolutional layers with 64 filters and 1 fully connected layer for the final classification. We only used one inner gradient step with 0.4 learning rate for all meta-curvature experiments for training and testing. The batch size was set to 32 (5-way) and 16 (20-way), and outer loop learning rate is 0.001 and we trained 60000 iterations.

\section{Few-shot classification on miniImagenet and tieredImagenet dataset}

\begin{table}[t]
\caption{Hyperparameters used for training MC2 on miniImagenet and tieredImagenet datasets with WRN-28-10 features and single hidden layer MLP.}
\label{table:miniimagenet_wrn_hyper}
\begin{center}
\begin{small}
\begin{tabular}{lccccccc}
\toprule
Hyperparameters & \multicolumn{4}{c}{miniImagenet} & \multicolumn{2}{c}{tieredImagenet} \\
\midrule
Features & \multicolumn{2}{c}{center} & \multicolumn{2}{c}{multiview} & \multicolumn{2}{c}{center} \\
& 1-shot & 5-shot & 1-shot & 5-shot & 1-shot & 5-shot \\
\midrule
Batch size & \multicolumn{6}{c}{16} \\
Total training iterations & \multicolumn{6}{c}{100,000} \\
Learning rate (inner loop) & \multicolumn{6}{c}{0.01} \\
\midrule
Learning rate (outer loop) & 0.0005 & 0.0003 & 0.0005 & 0.0003 & 0.00075 & 0.0001 \\
The number of inner steps  & 1 & 5 & 1 & 5 & 1 & 5 \\
The number of hidden units  & 1024 & 1024 & 512 & 4096 & 4096 & 4096 \\
Dropout rate over WRN-28-10 features & 0.5 & 0.3 & 0.5 & 0.7 & 0.5 & 0.5 \\
\bottomrule
\end{tabular}
\end{small}
\end{center}
\end{table}

\subsection{Baseline CNNs} For both 5-way 1-shot and 5-way 5-shot classification, we set the batch size 4 for 1 step experiments and 2 for 5 step experiments. 15 examples per class were used for evaluating the model after updates. In total, we ran 100,000 iterations for 1 step experiments and 200,000 iterations for 2 step experiments. The inner/outer learning rates are $\beta = 0.001, \alpha = 0.01$. We apply dropout rate 0.2 in the final linear layer for only MC1 and MC2 (other methods did perform worse with dropout). For cutout data augmentation, we cut out $36 \times 36$ random crops.

\subsection{WRN-28-10 features and MLP} We used the WRN-28-10 features provided by \cite{rusu-iclr19}. For miniImagenet, we provided the results from both center and multi-view features (average of center and corner crops). The dimension of feature was 640 and we used one hidden layer with ReLU activation function followed by a softmax classifier. We used separate meta-curvature matrices for each inner updates. The details of hyperparameters used for training MC2 is provided in Table \ref{table:miniimagenet_wrn_hyper}.

\section{Few-shot reinforcement learning}

The goal of few-shot learning in reinforcement learning (RL) is that an agent can quickly adapt to a new task with little prior experience. A distinct feature from the few-shot supervised learning task is that the RL objective is not generally differentiable. Therefore, we use policy gradient methods to estimate the gradient both for inner and outer loop gradients. In addition, policy gradient methods are generally on-policy, which means that the training data depends on the agent’s initial policy. Therefore, the initial policy (with the meta-learned initial parameters) needs to explore as diverse experiences as possible to get proper feedback from a new task. We described the method and interpretation with respect to supervised classification tasks, but it can be easily modified to RL setting. 

\subsection{Experimental setup}
We tested our method on complex high-dimensional locomotion tasks with the MuJoCo simulator \cite{mujoco}. Most of the settings are based on \cite{Finn-icml-2017} for fair comparison. We consider two simulated robots (HalfCheetah and Walker2d) and two types of task environments (to run in a forward/backward direction or a particular velocity). The network architecture is two hidden layers of size 100 with ReLU activations for both. We used the standard linear feature baseline estimator. We evaluated the performance after one policy gradient step with 20 trajectories. We compare against MAML-TRPO and MAML-PPO. In the original MAML, TRPO \cite{trpo} was used as the outer loop optimizer but we found out that using PPO \cite{ppo} consistently outperformed the TRPO. MAML-PPO is also computationally more efficient since MAML-TRPO requires third-order gradients (or computed by hessian-vector product instead). To the best of our knowledge, MAML-PPO has not been tested on this setup. We evaluated two variations of meta-curvature similar to the classification setup, MC1 and MC2, and used PPO as the meta-optimizer. Note that this is a preliminary result, so this is not by no means conclusive. We provide this information for the readers who might be interested in this direction.

\subsection{Experimental results}
Fig. \ref{fig:rl_experiment} shows the rewards obtained after one step policy gradient update. In the HalfCheetahDir experiment, our methods outperformed both strong baselines. MC1-PPO reached the same performance of a strong baseline, MAML-PPO three times faster. In HalfCheetahVel and Walker2dDir experiments, both MC2-PPO and MAML-PPO reached nearly the same performance, but in a more sample efficient manner. For Walker2dVel, MAML-TRPO showed the fastest convergence at the earlier meta-training stage, but our meta-curvatures outperformed eventually. In this setting, most of the rewards come from the survival reward (the agent gets $1.0$ reward for every step if they do not fall over). All methods were able to survive throughout the episode, but our methods run better at a given velocity. One thing we noticed that it stops obtaining more rewards and starts to degrade the performance in Walker2dDir experiment. The recently proposed approach \cite{promp} may alleviate this issue through better credit assignment in the meta-gradients. Combining it would be interesting direction to be explored.

\section{Visualization}
Fig. \ref{fig:meta_curvature_vis} is a visualization of meta-trained meta-curvature matrices for 5-way 1-shot classification task. To visualize the full matrix, $\B{M}_{mc}$, we picked up the matrices from the first convolutional layer in the small model (filter size 64). Therefore with the 3 color input channels, $\B{M}_f \in \mathbb{R}^{9 \times 9}$, $\B{M}_i  \in \mathbb{R}^{3 \times 3}$, $\B{M}_o  \in \mathbb{R}^{64 \times 64}$, and $\B{M}_{mc}  \in \mathbb{R}^{1728 \times 1728}$. The diagonal elements are high values, mostly $> 0.5$. Interestingly, there are also a lot of off-diagonal elements $> 0.5$ or $< -0.5$. Thus, they capture the dependencies between the gradients.

\end{document}